\documentclass[conference]{IEEEtran}
\IEEEoverridecommandlockouts
\usepackage{cite}
\usepackage{amsmath,amssymb,amsfonts}
\usepackage{algorithmic}
\usepackage{graphicx}
\usepackage{textcomp}
\usepackage{xcolor}
\usepackage{hanging}
\usepackage{graphicx}
\usepackage{booktabs}
\usepackage[utf8x,utf8]{inputenc}
\usepackage{apacite}

\usepackage[hyphens]{url}

\usepackage[T1]{fontenc}
\usepackage{CJKutf8}
\usepackage{fancyhdr}
\def\BibTeX{{\rm B\kern-.05em{\sc i\kern-.025em b}\kern-.08em
    T\kern-.1667em\lower.7ex\hbox{E}\kern-.125emX}}
\begin{document}

\title{Pseudo AI Bias\\
{\footnotesize \textsuperscript{*}Note: This is a preprint.}
\thanks{This study was partially funded by National Science Foundation (NSF) (Award 2101104 and 2100964) and National Academy of Education/Spencer Foundation. Any opinions, findings, conclusions, or recommendations expressed in this material are those of the author(s) and do not necessarily reflect the views of the founders. }
}

\author{\IEEEauthorblockN{1\textsuperscript{st} Xiaoming Zhai}
\IEEEauthorblockA{\textit{Department of Mathematics, Science, and Social Studies Education} \\
\textit{AI4STEM Education Center} \\
\textit{University of Georgia}\\
Athens, GA, United States \\
xiaoming.zhai@uga.edu}
\and
\IEEEauthorblockN{2\textsuperscript{nd} Joseph Krajcik}
\IEEEauthorblockA{\textit{CREATE for STEM Institute} \\
\textit{Michigan State University)}\\
East Lansing, MI, United States \\
krajcik@msu.edu}
}
\maketitle

\begin{abstract}
Pseudo Artificial Intelligence bias (PAIB) is broadly discussed in the literature, which can result in unnecessary AI fear in society, thereby exacerbating the enduring inequities and disparities in accessing and sharing the benefits of AI applications and wasting social capital invested in AI research. This study systematically reviews publications in the literature to present three types of PAIBs identified due to: a) misunderstandings, b) pseudo-mechanical bias, and c) over-expectations. We discussed the consequences of and solutions to PAIBs, including certifying users for AI applications to mitigate AI fears, providing customized user guidance for AI applications, and developing systematic approaches to monitor bias. We concluded that PAIB, due to misunderstandings, pseudo-mechanical bias, and over-expectations of algorithmic predictions, is socially harmful.
\end{abstract}


\section{Introduction}
As Artificial Intelligence (AI) is increasingly applied in solving problems in education, the potentially biased decisions that can be made by AI draw significant concerns. AI could assist human beings in scoring students’ constructed responses, analyzing processing data, and making administrative decisions \cite{rosenberg2022a, zhai2020a}. However, researchers realized that AI predictions might systematically deviate from the ground truth, generating results either in favor or against certain groups \cite{cheuk2021a,li2022a}. Such effects can be detrimental and result in social consequences that have drawn concerns about the use of AI since it was broadly applied in our everyday life \cite{kapur2021a,nazaretsky2021a}. 

While AI biases may result from deficit design, skewed training data, confounding of the algorithmic models, or the algorithm computational capacity, some are not. Defining AI bias and attributing bias outcomes to AI needs to be cautious. What if AI accurately predicts something unjust by nature? It can be even more complicated if attributions of biases to AI are artificially crafted, which has been seen in the literature \cite{zou2018a}. This may be particularly detrimental if understudied “biases” are arbitrarily attributed to AI only because users misunderstand the bias, inappropriately operate machine algorithms, misinterpret AI predictions, or have an over-expectation of AI predictions. To conceptualize and study these “biases,” we coined the term pseudo AI bias (PAIB). Attributing PAIB to AI is unfair and can disadvantage the potential of technologies in education. Moreover, overselling PAIB can result in unnecessary AI fear in society, exacerbate the enduring inequities and disparities in accessing and sharing the benefits of technologies \cite{rafalow2022a}, and waste social capital invested in AI research. 

In recognizing the potential harmful societal consequences of PIAB, this review study intends to conceptualize PIAB by reviewing the literature. We first review the definition of bias to identify the criteria for conceptualizing PAIBs. We then present three types of PAIBs identified in the literature and discuss the potential impacts of PAIBs on education and society. At last, we present some probable solutions. 

\section{Conceptualization of Pseudo AI Bias}
Bias was broadly referred to in social science and natural science, yet no consensus has been made on the definition. To identify the commonalities of bias across areas, we reviewed the definitions of bias in the literature in both areas, particularly focusing on the authoritative literature in each area (see Table 1). Our review yielded three critical properties of bias that are shared across areas: (a) deviation – bias measures the deviation between observations and ground truth (i.e., error); (b) systematic – bias refers to systematic error instead of random error; and (c) tendency -- bias is a tendency to favor or against some ideas or entity over others. These three properties characterize the idea of AI bias and are fundamental to uncovering PAIBs.

A review of the literature uncovers three types of PAIBs due to misunderstanding of bias, humans’ mis-operation of AI, or over-expectation of AI outcomes. Following we presented the three types of PAIBs.

\begin{table}[htbp]
\centering
\caption{Table 1 Definitions of bias in authoritative literature of social science and natural science}
\label{Table 1}
\resizebox{\columnwidth}{!}{%
\begin{tabular}{|p{2cm}|p{4.5cm}|}
\hline

Resources                       & Definition                                                               \\ 
\hline

Oxford English Dictionary       & Tendency to favor or dislike a person or thing, especially as a result of a preconceived opinion, partiality, prejudice. (Statistics) Distortion of a statistical result arising from the method of sampling, measurement, analysis, etc.                                                                               \\ 
\hline
Dictionary of Cognitive Science & A systematic tendency to take irrelevant factors into account while ignoring relevant factors (p. 352). Algorithms make use of additional knowledge (i.e., learning biases; p. 228).                                                                                                                                                                                                                                                \\ 
\hline
The Science Dictionary     & In common terms, bias is a preference for or against one idea, thing, or person. In scientific research, bias is described as any systematic deviation between the results of a study and the “truth.”                                                                                                                                                                                                                          \\ 
\hline
Wikipedia                       & Bias is disproportionate weight in favor of or against an idea or thing, usually in a way that is closed-minded, prejudicial, or unfair. Biases can be innate or learned. People may develop biases for or against an individual, a group, or a belief. In science and engineering, a bias is a systematic error. Statistical bias results from an unfair sampling of a population or from an estimation process that does not give accurate results on average (12/20/2021). 

\\ \hline
\end{tabular}%
}
\end{table}

\subsection{Category 1: Misunderstanding of Bias}

Error versus bias. Errors can be a bias only if they systematically happen. In scientific research, errors inevitably occur, as it is challenging to reach ground truth in measurement, prediction, classification, etc. \cite{brown2018a}. Therefore, instead of attempting to eliminate errors, researchers put efforts into reducing errors using various methods, such as refining measurement tools, averaging observations, or improving prediction algorithms \cite{kipnis2011a}. In contrast, the elimination of bias can occur once its root is identified and correct solutions are implemented \cite{mcnutt2016a}. Therefore, bias is more troublesome to humans than errors are, particularly when the bias is implicit \cite{warikoo2016a}. It is critical to correctly identify bias and eliminate it. Examples in the literature are broadly cited as evidence of AI bias and demonstrate AI’s harmfulness (see Cheuk, 2021). However, some of them may provide examples of AI prediction errors instead of bias.

Among such, the most popular is the example of humans being mistaken as Gorillas \cite{pulliam-moore2015a}. In a 2015 Twitter post, Jacky Alcine shared a screenshot in which a face-recognition algorithm tagged him and another African American as gorillas. This case shows evidence of the problem of algorithm prediction, indicating significant work needs to be done to improve prediction accuracy. However, although we find this example deplorable, numerous authors who cited this case as “AI bias” did not provide sufficient evidence that this terrible misidentification resulted from inherent AI bias—a single case is challenging to testify the errors “systematically” happen for a specific group. Another example includes face recognition software labeling a Taiwanese “blinking” (Rose, 2010), and an automated passport system not being able to detect whether an individual from Asia had his eyes open for a passport picture \cite{griffiths2016a}, etc. These cases reported AI errors caused by human programming of the algorithms, which need to be addressed through in-depth research, instead of AI bias because no evidence had shown that these errors occurred systematically when the cases were cited. 

Confusing bias with errors could extend the societal fear of AI, particularly toward certain groups. As mentioned above, we see these errors as deploring and we need to work hard to ensure these errors don’t happen, but in the world of programming, “bugs” happen, and we just need to make sure that public announcements such as bias are made with evidence. Currently, society invests trillions of dollars in technology developments every year, while certain groups underrepresented in society have been found less likely to benefit from these innovations \cite{rafalow2022a}. This consequence can be exacerbated by the fear of inaccurate technological outcomes that result from human errors, instead of some engrained bias in the technology. The unfairness could be even worse if the extended AI fear happens explicitly to the underrepresented groups. Therefore, researchers should use these mislabeled alerts to further examine potential bias due to programming and increase the accuracy of AI predictions instead of overselling errors as bias.

Favoriting percentage versus ground truth. Amazon's AI recruiting tool is a broadly cited case for evidence of "AI bias." According to \cite{dastin2018a}, this tool favored male applicants compared to females, resulting in the belief that the software was biased towards females and eventually disbanded by Amazon. Although widely citing this case to demonstrate that AI is biased against females, seldom has research referred to the prediction “deviations” from the ground truth—the outcomes based on recruiting criteria. A higher percentage favoring male candidate than females can hardly support a valid conclusion about AI bias before we know the actual numbers of applicants by gender and the false prediction cases. This PAIB results from a lack of information-- the gender breakdown of Amazon’s technical workforce, according to \cite{dastin2018a}, was not even disclosed at that time. Therefore, it is problematic to claim AI bias until more data becomes available and a deep investigation of gender parity can verify the PAIBs.

This type of PAIB harms people’s trust in AI. The predicted favoring percentage toward males is hard to justify discrimination before we know the “deviations.” As argued by \cite{howard2018a}, favoring a group might also be justified if the evidence supports the facts. For example, charging teenagers more for car insurance seems “discriminating” but arguably justified given that evidence shows a higher risk of accidents for teenage drivers. A higher charge for teenage drivers intends to prevent unnecessary risky driving by teenagers. A key to determining AI bias requires examining the ground truth and comparing it to the then findings instead of seeking the favoriting percentage. 

Fairness versus bias. Accuracy is the primary measure for algorithm predictions, while bias is the extent to which the results are systematically and purposely distorted. Therefore, computer scientists pursue high accuracy and avoid bias. Fairness, however, is beyond measurement. It is a social connotation incorporating factors such as access to social capital and the purpose of fairness \cite{teresi2013a}. A broadly cited example is Equality and Equity Fig. 1 Fairness: equality (left) and equity (right).(see Fig. 1).  When asked which picture represents the best practice of fairness, an educator might point to the right picture without hesitance as it represents the pursuit of equity—everyone has an opportunity to observe the game. Interestingly, when an Asian baseball player was shown this picture, he said he was never given equal opportunity to play in the game. Should athletes be granted equal opportunity? Why do observers deserve equity while athletes who play the game seldom have such considerations? These questions justify that fairness is based on social needs and the purpose of the activities. It is hard to absolutely declare that equity is closer to fairness than equality in every case. With such a complex societal concept, it is difficult to demand an algorithm to achieve fairness goals for every case. 

\begin{figure}[htbp]
\centerline{\includegraphics{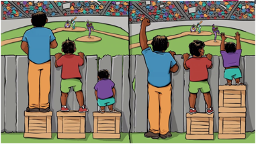}}
\caption{Fairness: equality (left) and equity (right).}
\label{Fig.1}
\end{figure}
Nevertheless, our research shows that AI was held full accountable for fairness in some cases. One example is the Correctional Offender Management Profiling for Alternative Sanctions (COMPAS), an algorithm used to predict the risk of recidivism, the tendency of a convicted criminal to re-offend. According to a report \cite{angwin2016a}, COMPAS achieved equal prediction accuracy for both black and white defendants. That is, a risk score that is correct in equal proportion for both groups. However, the report also suggests that black defendants were twice as likely to be incorrectly labeled as higher risk than white defendants. In reality, nevertheless, white defendants labeled as low risk were more likely to be charged with new offenses than black defendants with similar scores. According to research from four independent groups \cite{feller2016a}, to satisfy both accuracy and “fairness” is mathematically impossible given that Blacks were re-arrested at a much higher rate than Whites in reality. However, because the purpose of the risk score is to help the judge make decisions in court without considering the defendants' race, COMPAS’s performance satisfied the cornerstone criteria \cite{feller2016a}. This case illustrates how we have overly tasked AI algorithms for fairness, which could increase the suspicions of AI applications in societal issues. COMPAS yielded fairness concerns, instead of bias, but the scores predicted were equally accurate for the ethnic groups. Confusing bias with fairness is detrimental to the broad use of AI in solving problems. “Statistics and machine learning are not all-powerful \cite{kusner2020a}.” It is the users’ responsibility to clearly identify the societal goals of the prediction before using algorithm predictions. 

\subsection{Category 2: Pseudo-mechanical Bias}

In many cases, bias is created because of users’ ignorance of critical features of data or inappropriate operations in the algorithm application processes, instead of the capacity of algorithms. This type of bias is not due to the algorithm mechanism but human users’ operation and use of AI algirithms, termed pseudo-mechanical bias. It is feasible to avoid pseudo-mechanical bias as long as users appropriately apply the AI applications. Thus, attributing pseudo-mechanical bias to AI is misleading. Below, we introduced four categories of pseudo-mechanical bias presented in the literature (see Fig. 2).

Emergent error. Emergent error bias results from the out-of-sample input. It is widely assumed that AI algorithms can best predict samples close to the training data. In reality, this is not always the case, as ideal training data are challenging to collect. In this sense, users must be aware of the limitations of AI algorithms to avoid emergent bias. This type of error frequently appears in clinical science, where they are attributed to distributional shifts \cite{challen2019a}. That is, prior experience may not be adequate for new situations because of “out of sample” input no matter how experienced the “doctors” are. It is obvious that algorithms are especially weak in dealing with distributional shifts. Therefore, it is critical for users of AI systems to be aware of the limitations and apply AI to appropriate cases to avoid emergent errors.

Distinct types of data for training and testing. Currently, supervised machine learning requires using the same type of data for training and testing. Cheuk (2021) argues that algorithms would fail to respond to questions in one language if trained using a different language. In the example she raised, an AI trained in English was asked by a Chinese heritage “Have you eaten?” in Chinese (i.e., 
\begin{CJK}{UTF8}{gbsn} 
你吃了吗 
\end{CJK}). 
The author suspected that the algorithm would not be able to respond to this question as expected by the Chinese. Instead, the computer might respond intelligently, “No, I haven’t eaten,” because the computer might interpret the question as to whether the Chinese care about its diet. However, according to Chinese culture, the question “Have you eaten?” equals “How are you?” in English. This example is absolutely correct, but it is almost arbitrary for an English-speaking Chinese to ask AI questions in English (as the machine is trained in English as the author noted) and then expect the machine to answer the question following the Chinese culture. In other words, it is the users’ responsibility to be aware that the algorithm was trained in English and expect the response to follow the same conventions in English. These types of situations in which the algorithm is incapable of responding or interpreting could lead to PAIB, only because the AI algorithm was trained in a distinct way from what the users expected.

Distinct feature distribution among groups. In a recent study, Larrazabal and colleagues \cite{larrazabal2020a} examined the gender imbalance of algorithm capacity to diagnose disease based on imaging datasets. They found that for some diseases like Pneumothorax, the algorithms consistently performed better on males than females regardless of how researchers tried to improve the training set. This unintended error was created because patients have unique biological features that prevent algorithms from performing equally based on the gender of the patient.  For instance, females’ breasts occluded imagining of organs responsible for the disease when using x-ray technology to collect the imaging data of the diseased organs, resulting in poorer performance of algorithms when identifying disease. If users apply the algorithms without being aware of the limitations of the algorithms, it could generate PAIBs.

Disconnected interpretations of labels. In practice, if users interpret the meaning of the AI predictions differently from the information encapsulated in the training data, it could generate PAIBs. For example, Obermeyer and colleagues\cite{mcclure2018a} examined the most extensive commercial medical risk-prediction system, which is applied to more than 200 million people in the US each year. The system was trained using patients’ expenses and was supposed to predict the ideal solutions for high-risk patients. The authors found the “ideal solutions” discriminating-- Black patients with the same (predicted) risk scores as White patients tend to be sicker, resulting in unequal solutions provided to patients of color. In this case, the “predicted risk score” was generated based on past medical expenses, which is a combination of both the risk and the affordability of medical service. Suppose doctors interpret this score as an indicator of risk to providing medical care, it would likely put at risk a certain group’s life (e.g., Black patients who might generate less medical expense). This assumption is problematic because the assumed bias resulted from doctors’ misinterpretation of the risk score-- disregarding the fact that the scores reflect not only the severity of illness but also the affordability of the medical service provided to the patients. Obermeyer and colleagues\cite{obermeyer2019a} attributed the medical bias to label choice—fundamentally reflecting a disconnect between the interpretation of predictions (i.e., the risk of illness) and the information encapsulated in the training samples (i.e., both the risk of illness and the affordability of medical care). They also provided empirical evidence that altering the labels is effective in reducing biased results. 
\subsection{Category 3: Over Expectation}

In many cases, over-expectations can result in PAIBs. For example, Levin \citeyear{levin2016a} reported that in a beauty contest judged by Beauty.AI, 44 winners were nearly all white, with a limited number of Asian and dark skin individuals. Many pinpoint this case as evidence for AI bias. What makes it problematic is that the criteria for beauty are perceptional and vary across people--some care more about color while others concern more about symmetry. It is even challenging to generate a consensus criterion by humans. How could we expect AI to identify the beauty winners to satisfy a diverse population? In this case, AI should be treated as an intelligent individual, equal to a human individual, instead of some intelligence that can overcome diversity dilemmas. In other words, Beauty.AI learned from human-labeled examples and thus is expected to perform as what was learned. The bias does rest in the AI algorithms but in the humans that developed the algorithm. It may be unreasonable to expect the algorithm to perform in a way to satisfy all stakeholders, who might even disagree with each other.

\begin{figure*} [htbp]   
\centerline{\includegraphics[scale=0.7]{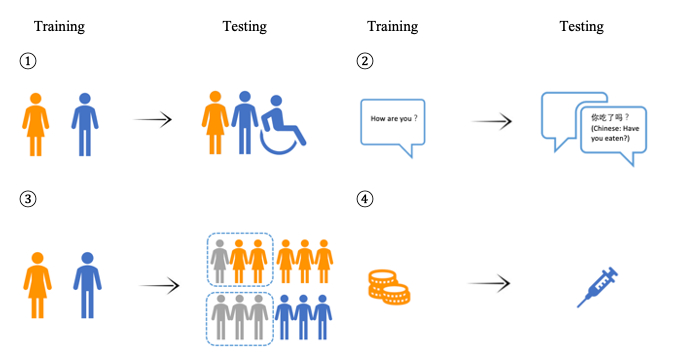}}
\caption{Fig. 2. Artificial mechanical bias. \textcircled{1} indicates emergent bias, where testing data include samples not included in the training data. \textcircled{2} indicates distinct data for training (e.g., Chinese vs. English). \textcircled{3} indicates Distinct samples for training and testing. Samples within the dashed box indicate having an illness, and gray indicates that the illness was predicted by the algorithm—\textcircled{4} indicates disconnected interpretations of labels.}
\label{Fig.2}
\end{figure*}

\section{Implications and Solutions}

This review by no means argues that AI biases are not worthy of being addressed. Indeed, “the growing scope of AI makes it particularly important to address [biased outcomes]” \cite{zou2018a}. In this review, however, we contend that understanding the mechanisms and potential bias of AI is essential for educational applications and other societal needs broadly. As the dual use of AI is increasingly recognized in the field, researchers have the responsibility to confirm the ethics before AI becomes mainstream in our society. In this review, our main concern focused on whether AI biases are correctly identified and appropriately attributed to AI. We focus this review on PAIBs as they have societally consequential \cite{zhai2021a,zhai2021b}. The public needs to realize that PAIBs result from users’ or developers’ misunderstanding, misoperation, or over-expectations, instead of the fault of the AI. Our position is that humans create algorithms and train machines, and the biases reflect human perspectives and human errors. Realizing the many kinds of PAIBs and how humans are at the center of them could help limit societal fear of AI.

PAIB adds to the existing and enduring human fear of AI. Such fear emerged earlier and faster than the progress in AI research because of numerous science fiction stories that struck and impacted human thinking about AI. The history of technophobia can be traced back to the 1920s when Rossum’s Universal Robots proposed scenes of enlarged robots (McClure, 2018). However, it did not draw much fear until the recent flood of AI developments, which refreshed humans’ fear of AI gleamed from these science fiction stories. Cave and Dihal analyzed 300 documents and identified four types of AI fears: inhumanity, obsolescence, alienation, and uprising \cite{cave2019a}. Because many AI fears are rooted in artificial scenes presented in science fiction books or movies \cite{bohannon2015a}, it is unsurprising that part of the fears might be artificial. Cave and Dihal’s \citeyear{cave2019a} survey of more than 1,000 UK citizens reveals that the common perception of AI is significantly more anxiety than excitement. This evidence, together with the findings from this review, exemplifies the importance of recognizing PAIBs.  

The societal consequences can even be worse if research enlarges the factors, such as PAIBs, that yield the fears. The users and developers of AI need to take responsibility to avoid PAIBs by clearly understanding the limitations of machine algorithms and the mechanisms of AI biases. Researchers, on the other side, have the responsibility to guide the use of AI applications and avoid extending PAIBs in academic work. Researchers need to put effort into examining and solving true AI biases instead of PAIBs. To deal with PAIBs, we recommend the following:

\subsection{Certify users for AI applications to mitigate AI fears}

PAIBs resulted because of users’ unfamiliarity with AI mechanics, as well as misunderstanding of bias that was enriched both by measurement and social connotations. To mitigate AI fears, users need professional education. Osborne, a researcher from the University of Oxford, suggests providing professional certifications to AI users such as doctors, judges, etc. \cite{gent2015a}. Such professional education could eliminate most PAIBs and help users mitigate AI fears.

\subsection{Provide customized user guidance for AI applications}

Some PAIBs developed due to the diverse configurations of the AI algorithms and the complex conditions of applying AI. For example, some algorithms might influence a certain group under a given condition. Unless users know of these constraints, they are not likely to appropriately employ the algorithms for predictions and decision makings. Therefore, customized user guidance, accompanied by professional education, is essential to alleviate the fear of AI and use AI appropriately. 

\subsection{Develop systematic approaches to monitor and reducing bias}

The best approach to clarifying PAIBs will stem from the development of tools and criteria to systematically monitor bias in algorithms. Such an approach could ease the public worry about AI applications and prevent biases in the first place. Aligned with this goal, the Center for Data Science and Public Policy created the Aequitas, an open-source bias toolkit to audit machine learning algorithm biases \cite{saleiro2020a}. This toolkit could audit two types of biases: 1) biased actions or interventions that are not allocated in a way that is representative of the population and 2) biased outcomes through actions or interventions that result from the system being wrong about certain groups of people. Besides, methodological efforts are needed to reduce biased results, which can avoid PAIBs. For example, Li, Xing, and Leite \citeyear{li2022a} developed a fair logistic regression algorithm and achieved less biased results. Once true AI bias is reduced significantly, the threaten of PAIBs will be limited. 

AI has the potential to serve as a viable tool and partner for many professions, including education.  However, AI will never reach its potential unless researchers eliminate sources of bias, and the public comes to know that many biases result from human errors in operating algorithms and in misinterpreting the results of AI. Although dealing with AI bias is critical and should draw substantial attention, overselling PAIBs can be detrimental, particularly to those who have fewer opportunities to learn AI, and may exacerbate the enduring inequities and disparities in accessing and sharing the benefits of AI applications.

\bibliographystyle{apacite}
\bibliography{references}

\vspace{12pt}

\end{document}